\PassOptionsToPackage{table}{xcolor}
\documentclass{article}
\usepackage{iclr2027_conference,times}
\usepackage[T1]{fontenc}
\usepackage{amsmath,amssymb,amsthm,bm}
\usepackage{graphicx,booktabs,array}
\usepackage{algorithm,algpseudocode}
\usepackage[table]{xcolor}
\usepackage{enumitem}
\usepackage{hyperref}
\usepackage{xurl}
\hypersetup{colorlinks=true,linkcolor=blue!45!black,citecolor=blue!45!black,urlcolor=blue!45!black,
 pdfauthor={Suxin Ji, Hungtao Wan, Mingjun Liu, An Zhang}}
\newcommand{\base}{\theta_0}
\newcommand{\R}{\mathbb{R}}

\newcommand{\norm}[1]{\left\lVert #1\right\rVert}
\definecolor{gaincolor}{HTML}{00796B}
\definecolor{losscolor}{HTML}{B2473E}
\definecolor{tablehead}{HTML}{E8EFF3}
\definecolor{tablerow}{HTML}{F4F6F8}
\newcommand{\gain}[1]{\textcolor{gaincolor}{$\uparrow\,#1$}}

\newcommand{\updelta}[1]{\textcolor{gaincolor}{\raisebox{-0.45ex}[0pt][0pt]{\scriptsize$\uparrow\!#1$}}}
\newcommand{\downdelta}[1]{\textcolor{losscolor}{\raisebox{-0.45ex}[0pt][0pt]{\scriptsize$\downarrow\!#1$}}}

\title{When Are Nearby Experts Composable?\\Cross-Task Model Merging\\in Neural Thickets}
\author{Suxin Ji$^{1}$\quad Hungtao Wan$^{1}$\quad Mingjun Liu$^{2}$\quad An Zhang$^{3}$ \\
$^{1}$Independent Researcher \\
$^{2}$University of Electronic Science and Technology of China\quad $^{3}$Chongqing University}
\iclrfinalcopy
\begin{document}
\maketitle
\lhead{}\renewcommand{\headrulewidth}{0pt}
\begin{abstract}
Random perturbations around a shared checkpoint can produce experts for distinct tasks, but shared origin does not ensure that their gains survive composition. We formulate expert-pair selection for joint base-relative retention, measured by strict joint success and worst-task gain with unconditional abstention accounting. An exact finite-difference decomposition and an arbitrarily small orthogonal counterexample show why proximity and orthogonality alone are insufficient. We instantiate this formulation with quality-constrained spectral selection, a fixed-budget framework combining cross-task screening, layerwise pair geometry, separate validation, and Base fallback. Across six checkpoints, 21 task pairs, and five independent perturbation pools, the procedure improves strict joint success over Top-1 averaging throughout and leads the four-method model-scale comparison on four of six checkpoints. For the five models with at least 1.5B parameters, matched quality-only, global-cosine, and spectral ranking attain $49.1\%$, $50.5\%$, and $53.1\%$ joint success. Under this matched protocol, the spectral selector also yields the highest worst-task gain and Weak Pareto rate and the lowest regression rate. Mean worst-task gain reaches $+0.61$ percentage points, while $17.1\%$ regression and $13.0\%$ abstention show that composability remains conditional.
Project page: \url{https://anonymous.4open.science/r/Thickets-Merge/}.
\end{abstract}

\section{Introduction}
\label{sec:introduction}
A useful merge must retain multiple capabilities: improving programming while degrading mathematics can raise an average score without achieving joint retention. Methods include geometric averaging in Model Stock \citep{model_stock}, task arithmetic in MetaGPT \citep{metagpt}, sparse updates in DARE and Model Breadcrumbs \citep{dare,model_breadcrumbs}, adaptive coefficients in AdaMerging \citep{adamerging}, and task-specific masks and rescalers in EMR-Merging \citep{emr_merging}. MergeBench examines retention tradeoffs in domain-specialized LLMs \citep{mergebench}, motivating taskwise assessment of whether one merged checkpoint retains the distinct capabilities supplied by its constituent experts.

Random experts make this question concrete. Neural Thickets identifies task-improving models near a pretrained checkpoint through RandOpt's sample--select--ensemble procedure \citep{neural_thickets}. SubZero and LOZO instead use structured random perturbations to estimate iterative optimization updates \citep{subzero,lozo}; Iterative RandOpt adds repeated sampling and distillation \citep{iterative_randopt}. We study composition of separately selected task experts from a shared candidate population. A shared checkpoint provides a common parameterization for combining updates, but does not establish that their gains survive addition.

Geometric structure offers useful signals. Task Singular Vectors addresses task-subspace interference \citep{tsv}, KnOTS aligns low-rank updates \citep{knots}, isotropic merging separates common and task-specific subspaces \citep{isotropic_merging}, and Essential Subspace Merging connects update selection to output shifts \citep{essential_subspace}. Functional evaluation remains essential: individually beneficial updates can interact adversely or damage another task. We study \emph{joint retention under a specified expert-selection and merging procedure}.

The reference and objective matter. Pareto-oriented coefficient search examines tradeoffs among capabilities \citep{recycle_merging}; beating selected specialists or increasing an average need not improve both tasks over Base. We distinguish strict joint success, Weak Pareto improvement, and worst-task gain, counting abstentions in the unconditional evaluation. An exact finite-difference identity separates individual effects from interactions, and an arbitrarily small orthogonal counterexample establishes a geometric non-guarantee. Composability is therefore evaluated for a specified selection and merging procedure.

We instantiate this formulation with quality-constrained spectral selection, a fixed-budget expert-pair selection framework: cross-task scores screen quality, layerwise subspace overlap ranks pairs, and separate validation calibrates merges or returns Base. Subspace-guided expert selection \citep{subspace_boosting}, rewarded-perturbation consolidation \citep{corp}, and evolutionary merging recipes \citep{evolutionary_merging} provide complementary precedents. Matched quality-only and global-cosine controls assess pair ranking while keeping candidate screening, coefficient calibration, validation budget, and decision rule fixed throughout the comparison.

We evaluate six instruction-tuned Qwen2.5 \citep{qwen25}, Llama 3.1 \citep{llama3}, and OLMo 3 \citep{olmo3} checkpoints across 21 task pairs and five independent pools. Ours exceeds Top-1 throughout and leads the four-method comparison on four checkpoints; Direct Search leads Qwen 7B and TIES leads OLMo 3. For the five $\geq$1.5B models, strict joint success is $53.1\%$ (quality-only: $49.1\%$; cosine: $50.5\%$), and mean worst-task gain is $+0.61$ pp.

Our central contribution is joint-retention pair selection for local perturbation experts: deciding which independently selected experts around a shared checkpoint can be composed to improve both tasks over Base.
\begin{itemize}[leftmargin=*,itemsep=2pt,topsep=3pt]
\item \textbf{Problem formulation.} We formulate cross-task composability as joint base-relative retention rather than average improvement or specialist-relative performance, with unconditional accounting of abstentions.
\item \textbf{Failure characterization.} We characterize why local proximity and simple orthogonality are insufficient through an exact interaction decomposition, an arbitrarily small orthogonal counterexample, and radius identities.
\item \textbf{Decision procedure.} We introduce bounded expert-pair selection combining functional screening, layerwise subspace ranking, coefficient calibration, and validation with abstention. Evaluation spans six checkpoints, seven tasks, 21 pairs, and five independent perturbation pools, with matched controls and unconditional outcome accounting.
\end{itemize}

\section{Related Work}
\label{sec:related}
\paragraph{Random search and expert consolidation.}
Evolution strategies, MeZO, and Black-Box Tuning use random perturbations for optimization \citep{evolution_strategies,mezo,black_box_tuning}. We instead compose separately selected perturbation experts. RandOpt ensembles local experts; Iterative RandOpt distills them \citep{neural_thickets,iterative_randopt}. CoRP combines reward/compatibility weighting, validation, and population-informed search \citep{corp}; evolutionary and Pareto methods search checkpoint combinations \citep{evolutionary_merging,recycle_merging}. Our target is joint base-relative retention across tasks.

\paragraph{Weight merging and merge geometry.}
Checkpoint averaging \citep{model_soups}, task arithmetic \citep{task_arithmetic}, and importance/statistics-based rules \citep{fisher_merging,regmean} combine given updates. Comparing random and fine-tuned experts additionally requires matched quality, displacement, and cost. Connectivity and transfer motivate shared-base composition \citep{frankle_linear_connectivity,neyshabur_transfer}; Git Re-Basin and tangent-space arithmetic address alignment and functional disentanglement \citep{git_rebasin,tangent_space}. Task-vector and subspace methods provide geometric signals \citep{tsv,essential_subspace}.

\paragraph{Mergeability and merge selection.}
\emph{Will it Merge?} quantifies fine-tuned-model mergeability and links it to base knowledge \citep{will_it_merge}. \emph{Demystifying Mergeability} studies method- and partner-dependent predictors for fine-tuned CLIP pairs, including left/right task-vector singular-subspace overlap \citep{demystifying_mergeability}. SimMerge uses functional/structural similarity to select checkpoint subsets, operators, and order \citep{simmerge}. CABS+ combines conflict-aware sparsification and adaptive weights; its Relative Synergy Score guides model selection \citep{cabs_plus}. \citet{demystifying_mergeability} report positive predictive associations for leading right-subspace alignment; our ranking favors lower overlap after functional screening. We therefore interpret overlap direction as protocol-dependent. Prediction assesses a given merge; our decision jointly specifies separately selected local perturbation experts from one shared population, strict improvement on both tasks over Base, and pair selection under bounded validation with explicit abstention.

\section{From Expert Quality to Composability}
\label{sec:setup}
\paragraph{Candidates and selected experts.}
Let $\base\in\R^d$ be a shared checkpoint and $\theta_i=\base+\delta_i$ a sampled candidate. For task $t$, $S_t(\theta;D)$ is a higher-is-better score on data $D$. Selection data $D_t^{\mathrm{sel}}$ determine a task-specific pool $\mathcal{E}_t$ and Top-1 index $e_t$; separate validation data calibrate merges, and held-out data determine final gains. The shared checkpoint may already be instruction tuned. Candidates are sampled around the shared checkpoint and subsequently identified by their resulting parameter updates and task scores; selection and merging operate on these realized candidates rather than on seed identifiers alone.

\paragraph{Merging and reference models.}
For tasks $A,B$, arithmetic merging gives
\begin{equation}
\theta_{ij}(\alpha,\beta)=\base+\alpha\delta_i+\beta\delta_j,
\qquad i\in\mathcal{E}_A,\ j\in\mathcal{E}_B.
\label{eq:merge}
\end{equation}
Top-1 averaging uses $\alpha=\beta=1/2$; calibration selects coefficients on validation data. Deployment uses one checkpoint and one response per example. On held-out data, define
\begin{equation}
g_t(\theta)=S_t(\theta)-S_t(\base),\qquad
w(\theta)=\min\{g_A(\theta),g_B(\theta)\}.
\label{eq:gain}
\end{equation}
For accuracy, gains are percentage points (pp). Strict joint success (SJ) requires $w>0$. Weak Pareto improvement requires $w\geq0$ and $\max(g_A,g_B)>0$; a two-task tie does not qualify. Regression means $w<0$. Separately, \emph{specialist dominance} compares each task with its selected expert rather than Base, and can hold even when a task remains below Base.

\paragraph{Unconditional reporting.}
SJ is the percentage of successful pair--pool evaluations. Mean worst-task gain (WG) averages $w$; mean gain averages $(g_A+g_B)/2$. If no merge passes validation, the procedure returns Base and records abstention. Such outcomes remain in every denominator and contribute zero to gain averages, but are neither joint successes, Weak Pareto improvements, nor regressions. Validation acceptance and held-out success are distinct events.

\paragraph{Conditional composability.}
For a fixed candidate/selection distribution $Q$, merge rule $M$, and margins $m_t\geq0$, define
\begin{equation}
\kappa(Q,M,\mathbf m)=\Pr_{z\sim Q}\!\left[\forall t,\;
g_t\bigl(M(z)\bigr)>m_t\right].
\label{eq:kappa}
\end{equation}
Here, $z$ denotes a sampled selection instance under $Q$.
Our finite evaluation measures protocol-specific frequency over selected tasks and pools. It does not estimate the volume density of successful models in weight space or establish that a different pairing rule has the same success rate.

\section{Why Nearby Experts Can Conflict}
\label{sec:analysis}
\paragraph{Merging changes direction and radius.}
For $u=\alpha\delta_i$ and $v=\beta\delta_j$,
\begin{equation}
\norm{u+v}^{2}=\alpha^2\norm{\delta_i}^{2}+\beta^2\norm{\delta_j}^{2}
+2\alpha\beta\langle\delta_i,\delta_j\rangle.
\label{eq:radius}
\end{equation}
Averaging two orthogonal updates of norm $r$ gives radius $r/\sqrt2$, whereas adding them gives $\sqrt2r$. Fixed coefficients therefore do not isolate directional compatibility. Global norm matching also does not equalize layerwise sensitivity.

\paragraph{Orthogonality does not remove cross-task effects.}
Near-zero cosine in high dimensions does not imply that two updates preserve each other's capabilities. For normalized vectors with cosine $c$, the two-vector Gram eigenvalues are $1\pm|c|$: this spectrum re-expresses absolute cosine. Matrix-level singular subspaces carry different information, but their predictive value still needs controlled evaluation.

\paragraph{An exact decomposition separates effects from interactions.}
For a smooth diagnostic loss, let $\Delta L_t(u)=L_t(\base+u)-L_t(\base)$ and
\begin{equation}
I_t(u,v)=L_t(\base+u+v)-L_t(\base+u)-L_t(\base+v)+L_t(\base).
\label{eq:interaction}
\end{equation}
Then $\Delta L_t(u+v)=\Delta L_t(u)+\Delta L_t(v)+I_t(u,v)$. Even $I_t=0$ cannot ensure improvement if an update's cross-task harm outweighs the other's benefit. Appendix~\ref{app:theory} derives the Hessian-integral form and an arbitrarily small orthogonal counterexample. These are statements about smooth losses, not guarantees for discrete LLM accuracy. The analysis motivates cross-task screening and validation of actual merges, without deriving the spectral score used to prioritize pairs.

\section{Selecting and Merging Local Experts}
\label{sec:method}

\begin{figure}[!htbp]
\centering
\includegraphics[width=\linewidth]{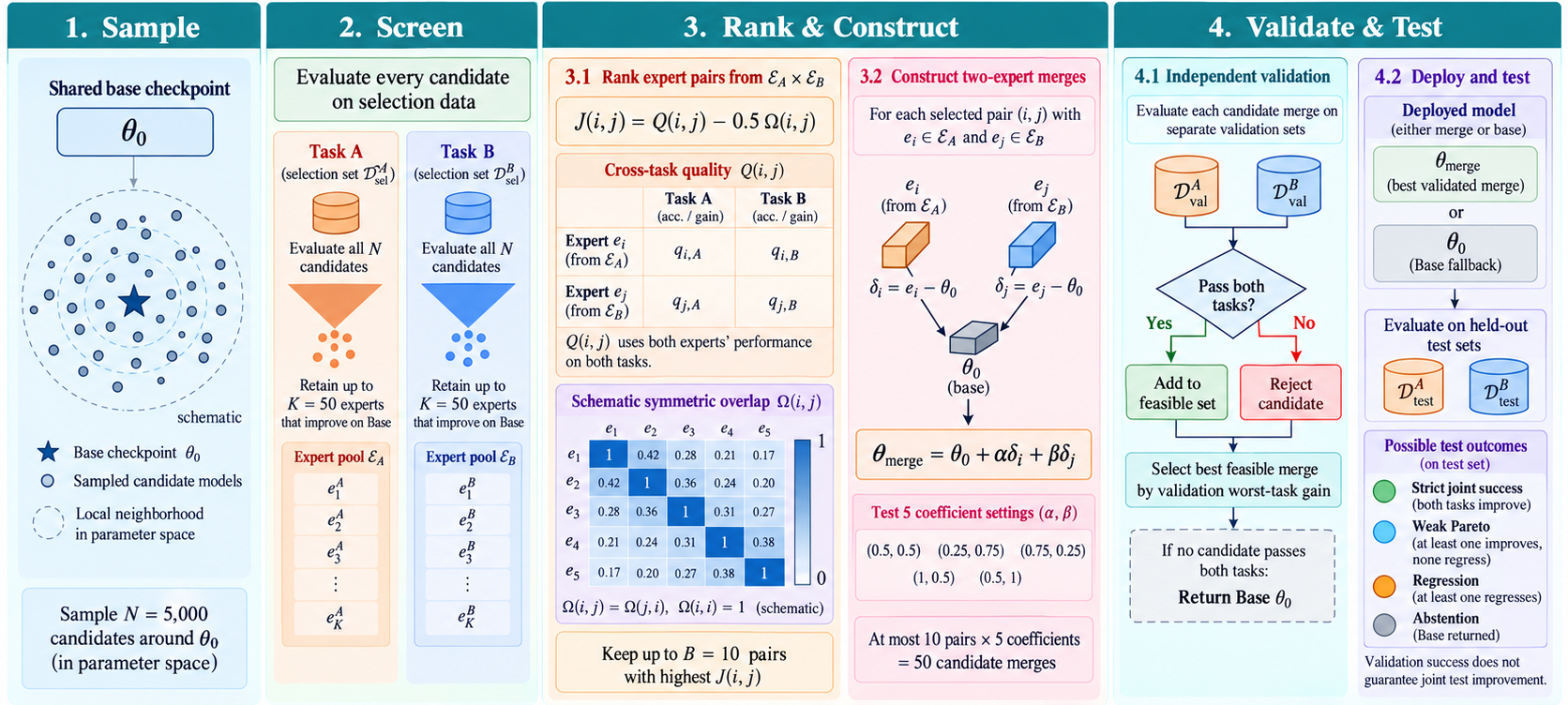}
\caption{\textbf{From a shared checkpoint to a validated two-expert merge.} Screen candidates, rank distinct expert pairs, and validate at most $10\times5=50$ merges. Screening permits ties with Base; validation requires both gains $>0$. Quality entries denote base-relative selection gains. Neighborhood and overlap are schematic; weak Pareto includes strict joint success. The best feasible merge or Base proceeds to held-out testing, without guaranteed improvement.}
\label{fig:method}
\end{figure}

Given local candidates for two tasks, the decision is which pair to compose under a fixed evaluation budget. Layerwise overlap is a ranking signal within functional screening and validation, rather than a standalone definition of mergeability. We instantiate this joint-retention framework with \emph{quality-constrained spectral selection}: functional screening, geometric pair prioritization, and bounded validation with fallback (Figure~\ref{fig:method}; Algorithm~\ref{alg:selection}, Appendix~\ref{app:algorithm}). Both experts are scored on both tasks; selection, validation, and testing use disjoint data. Building on task-vector and subspace merging \citep{task_arithmetic,tsv,subspace_boosting}, the procedure uses no gradient updates.

\paragraph{Stage 1: screen task-conditioned pools.}
Score every candidate on every target task using $D_t^{\mathrm{sel}}$, obtaining base-relative gains $g_t^{\mathrm{sel}}(\theta_i)$. Retain up to $K=50$ candidates per task satisfying $g_t^{\mathrm{sel}}(\theta_i)\geq\eta_t$, with $\eta_t=0$ pp. Require $i\neq j$ for a two-expert merge. Direct Search is the single-candidate multitask baseline: among candidates scored on both tasks, it selects one realized candidate rather than composing two distinct experts. For two-expert selection, an empty feasible pair set triggers base fallback, recorded as abstention. Screening uses no final-test information and does not establish generalization.

\paragraph{Stage 2: rank pairs by quality and subspace overlap.}
For an eligible two-dimensional parameter tensor $\ell$, write the rank-$r_\ell$ singular subspaces of its update as
\begin{equation}
\Delta W_{i,\ell}\approx U_{i,\ell}\Sigma_{i,\ell}V_{i,\ell}^{\top},\qquad
\omega_\ell(i,j)=\frac{\norm{U_{i,\ell}^{\top}U_{j,\ell}}_F^2+
\norm{V_{i,\ell}^{\top}V_{j,\ell}}_F^2}{2r_\ell}.
\label{eq:overlap}
\end{equation}
We use orthonormal columns and a common positive rank at each tensor, with the rank capped at eight and the weighting scheme fixed across all primary runs. Then $0\leq\omega_\ell\leq1$. Aggregate as $\Omega(i,j)=\sum_\ell w_\ell\omega_\ell(i,j)$, where $w_\ell\geq0$ and $\sum_\ell w_\ell=1$. We allocate half the total weight to attention matrices and half to MLP matrices, uniformly within each family, after excluding ineligible tensors. Non-matrix parameters remain in the merge but do not contribute to the overlap score. Sign-invariant subspace overlap discards singular-value magnitudes and does not measure task loss.

Rank pairs using
\begin{equation}
J(i,j)=\min_{t\in\{A,B\}}
\frac{g_t^{\mathrm{sel}}(\theta_i)+g_t^{\mathrm{sel}}(\theta_j)}{2}
-\lambda\Omega(i,j),\qquad\lambda\geq0.
\label{eq:ranking}
\end{equation}
The first term penalizes observed cross-task harm without assuming accuracy is linear in weights. The second favors lower subspace overlap at comparable score quality. We use $\lambda=0.5$ across all models, pairs, and pools. Setting $\lambda=0$ gives quality-only ranking; Global Cosine substitutes absolute full-update cosine for $\Omega$, with all other settings fixed (Table~\ref{tab:methods}).

\paragraph{Stage 3: calibrate within a fixed budget.}
Evaluate up to $B$ top-ranked distinct pairs on $\mathcal{A}=\{(0.5,0.5),(0.25,0.75),(0.75,0.25),(1,0.5),(0.5,1)\}$. Unlike AdaMerging's unsupervised coefficient optimization \citep{adamerging}, this gate uses separate labeled validation data. A feasible merge must satisfy $g_A^{\mathrm{val}}>0$ and $g_B^{\mathrm{val}}>0$. Conditional composability permits abstention: select the largest worst-task gain, breaking ties by mean gain and fixed candidate/coefficient order, or return Base if no proposal is feasible. We use $K=50$, $B=10$, five coefficient proposals per pair, and a maximum of 50 validated merges per task pair (Appendix~\ref{app:protocol}).

\paragraph{Cost and scope.}
Screening costs $NT$ task evaluations; calibration adds at most $2B|\mathcal{A}|$ per task pair. Across seven tasks, the default uses 35,000 screening and at most 2,100 calibration evaluations per pool, excluding base evaluation, SVD, and reconstruction overhead. These operation counts do not establish matched wall-clock cost. The deployed model has fixed weights, unlike input-dependent WEMoE \citep{wemoe}; TIES is a separate interference-handling comparator \citep{ties}. Composing more than two experts requires joint evaluation: pairwise compatibility does not imply compatibility of their sum.

\section{Experiments}
\label{sec:experiments}
We address four research questions:
\begin{itemize}[leftmargin=3.2em,labelwidth=2.8em,labelsep=0.4em,itemsep=0pt,topsep=2pt]
\item[\textbf{RQ1:}] How does joint improvement vary across models and scales?
\item[\textbf{RQ2:}] How does the full method compare with alternative search and merge rules?
\item[\textbf{RQ3:}] Does spectral pair ranking add value under a matched pipeline?
\item[\textbf{RQ4:}] How much does performance vary across independent perturbation pools?
\end{itemize}

\subsection{Evaluation setup}
We use Qwen2.5-Instruct at 0.5B, 1.5B, 3B, and 7B, Llama-3.1-8B-Instruct, and OLMo 3-7B-Instruct \citep{qwen25,llama3,olmo3}. The seven tasks are Countdown, GSM8K, MATH-500, OlympiadBench, MBPP, ROCStories, and USPTO, covering all 21 unordered pairs \citep{neural_thickets,gsm8k,olympiadbench,mbpp,rocstories}. Each model has five independently sampled pools, giving 105 pair--pool evaluations. Table~\ref{tab:methods} reports the five-model aggregate with at least 1.5B parameters (525 evaluations); Table~\ref{tab:results} includes all six checkpoints, with 0.5B as a separate stress test.

We sample $N=5{,}000$ candidates per pool, retain up to $K=50$ per task, and validate up to ten pairs with five coefficients each. Selection and validation use disjoint sets of 200 examples per task; final assessment uses held-out test data. MATH-500 remains held out, with selection and validation drawn from MATH training data. We fix $\lambda=0.5$, screening threshold $\eta_t=0$, and a strictly positive validation-gain gate across all models. Inference uses BF16, greedy decoding, and one response per item. Appendix~\ref{app:protocol} summarizes the primary evaluation configuration and aggregation rules.

\subsection{Model-scale composability (RQ1)}
\label{sec:rq1}
\begin{table}[t]
\centering\footnotesize
\setlength{\tabcolsep}{4pt}\renewcommand{\arraystretch}{1.10}
\caption{\textbf{Cross-task composability across model scales.} Each model covers 21 task pairs and five independent pools (105 outcomes). SJ: strict joint success (\%); WG: mean worst-task gain (pp). Bold marks metric-wise row bests. Small arrow values give Ours minus the best of Top-1, TIES, and Direct for each model and metric (pp), computed from displayed entries.}
\label{tab:results}
\begin{tabular}{>{\raggedright\arraybackslash}p{0.17\linewidth}*{2}{*{3}{>{\raggedleft\arraybackslash}p{\dimexpr(\linewidth-0.17\linewidth-0.25\linewidth-18\tabcolsep)/6\relax}}>{\raggedleft\arraybackslash}p{0.125\linewidth}}}\toprule
\rowcolor{tablehead}&\multicolumn{4}{c}{SJ $\uparrow$ (\%)}&\multicolumn{4}{c}{WG $\uparrow$ (pp)}\\
\cmidrule(lr){2-5}\cmidrule(l){6-9}
\rowcolor{tablehead}Model&Top-1&TIES&Direct&Ours&Top-1&TIES&Direct&Ours\\\midrule
Qwen2.5-0.5B & 11.4 & 23.8 & 21.9 & \textbf{25.7}\updelta{1.9} & -1.02 & -0.28 & -0.12 & \textbf{-0.06}\updelta{0.06}\\
\rowcolor{tablerow} Qwen2.5-1.5B & 18.1 & 33.3 & 38.1 & \textbf{42.9}\updelta{4.8} & -0.71 & +0.15 & +0.23 & \textbf{+0.42}\updelta{0.19}\\
Qwen2.5-3B & 26.7 & 44.8 & 49.5 & \textbf{55.2}\updelta{5.7} & -0.34 & +0.31 & +0.43 & \textbf{+0.69}\updelta{0.26}\\
\rowcolor{tablerow} Qwen2.5-7B & 33.3 & 52.4 & \textbf{61.0} & 59.0\downdelta{2.0} & -0.12 & +0.48 & \textbf{+0.77} & +0.72\downdelta{0.05}\\
Llama-3.1-8B & 28.6 & 47.6 & 50.5 & \textbf{57.1}\updelta{6.6} & -0.22 & +0.36 & +0.44 & \textbf{+0.65}\updelta{0.21}\\
\rowcolor{tablerow} OLMo 3-7B & 27.6 & \textbf{53.3} & 49.5 & 51.4\downdelta{1.9} & -0.26 & \textbf{+0.61} & +0.46 & +0.58\downdelta{0.03}\\
\bottomrule\end{tabular}\end{table}

\begin{figure}[t]
\centering
\includegraphics[width=\linewidth]{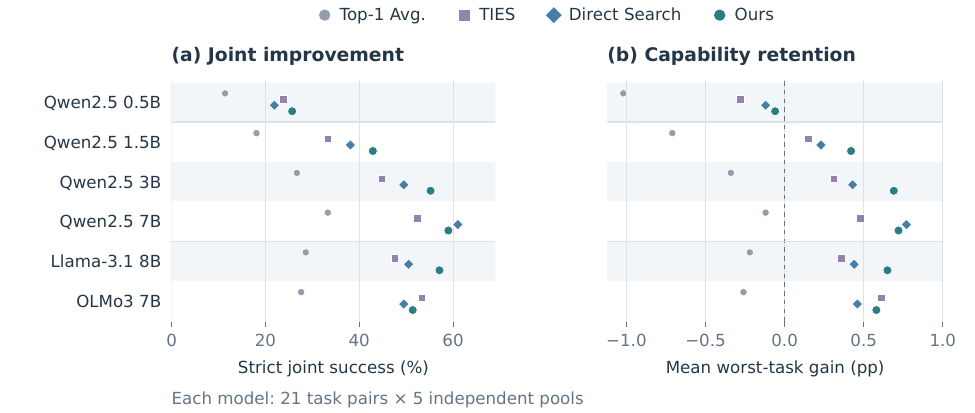}
\caption{\textbf{More joint successes do not guarantee positive worst-task gain.} Model-wise SJ and WG for four methods. Each row summarizes 21 task pairs and five pools. Our method improves on Top-1 throughout, but Direct Search leads Qwen 7B and TIES leads OLMo 3. No uncertainty intervals are inferred from these model-level aggregates.}
\label{fig:scale}
\end{figure}
\paragraph{The full method leads four of six models, with meaningful exceptions.}
Our SJ exceeds Top-1 across all six checkpoints (Table~\ref{tab:results}). Within Qwen, it increases $25.7\rightarrow42.9\rightarrow55.2\rightarrow59.0\%$ with model size. Direct Search nevertheless leads at 7B ($61.0>59.0$), and TIES leads OLMo 3 ($53.3>51.4$). At 0.5B, our WG remains $-0.06$ pp despite improved SJ. Thus success frequency, average worst-task retention, and the best comparison method need separate assessment (Figure~\ref{fig:scale}).

\subsection{Primary comparison and abstention (RQ2)}
\label{sec:rq2}
We separate system-level and component-level evidence. Top-1, TIES, and Direct Search compare complete merge/search procedures, whereas Quality-only, Global Cosine, and Ours hold candidate pools, screening, coefficient proposals, validation budget, and the decision gate fixed, isolating the pair-ranking criterion.
\begin{table}[t]
\centering\footnotesize
\setlength{\tabcolsep}{3.5pt}\renewcommand{\arraystretch}{1.12}
\caption{\textbf{Aggregate comparison for the five $\geq$1.5B models.} 525 pair--pool evaluations per method; all metrics include abstentions. Rates are in \%; gains are in pp. Bold marks columnwise optima. Small arrow values, omitted for Abstain, give Ours minus the best comparator per metric (pp), from displayed entries; green/red denotes better/worse.}
\label{tab:methods}
\begin{tabular}{>{\raggedright\arraybackslash}p{0.24\linewidth}*{4}{>{\raggedleft\arraybackslash}p{\dimexpr(\linewidth-0.24\linewidth-0.125\linewidth-14\tabcolsep)/5\relax}}>{\raggedleft\arraybackslash}p{0.125\linewidth}>{\raggedleft\arraybackslash}p{\dimexpr(\linewidth-0.24\linewidth-0.125\linewidth-14\tabcolsep)/5\relax}}\toprule
\rowcolor{tablehead}Method&SJ $\uparrow$&WG $\uparrow$&\shortstack{Weak\\Pareto $\uparrow$}&\shortstack{Mean\\gain $\uparrow$}&\mbox{Regression $\downarrow$}&Abstain $\downarrow$\\\midrule
Top-1 Avg. & 26.9 & -0.33 & 42.1 & +0.36 & 57.9 & \textbf{0.0}\\
\rowcolor{tablerow} TIES & 46.3 & +0.38 & 61.3 & +1.09 & 38.7 & \textbf{0.0}\\
Direct Search & 49.7 & +0.47 & 64.6 & +1.28 & 35.4 & \textbf{0.0}\\
\rowcolor{tablerow} Quality-only, $\lambda=0$ & 49.1 & +0.49 & 66.1 & +1.39 & 20.4 & 13.5\\
Global Cosine & 50.5 & +0.54 & 67.6 & +1.48 & 19.0 & 13.3\\
\rowcolor{tablehead} \textbf{Ours} & \textbf{53.1}\updelta{2.6} & \textbf{+0.61}\updelta{0.07} & \textbf{69.9}\updelta{2.3} & \textbf{+1.57}\updelta{0.09} & \textbf{17.1}\textcolor{gaincolor}{\raisebox{-0.45ex}[0pt][0pt]{\scriptsize$\downarrow\!1.9$}} & 13.0\\
\bottomrule\end{tabular}
\par\vspace{3pt}\begin{minipage}{\linewidth}\footnotesize Quality-only and Global Cosine retain the full screening, validation budget, and gate, changing only pair ranking. Weak Pareto includes SJ; returning Base is neither improvement nor regression.\end{minipage}\end{table}

\paragraph{Joint retention improves beyond positive average gain.}
Our SJ is $53.1\%$, versus $26.9\%$, $46.3\%$, and $49.7\%$ for Top-1, TIES, and Direct Search (Table~\ref{tab:methods}). All six methods have positive mean gain, yet Top-1's WG is $-0.33$ pp: averages can conceal cross-task damage.

\begin{figure}[t]
\centering
\includegraphics[width=\linewidth]{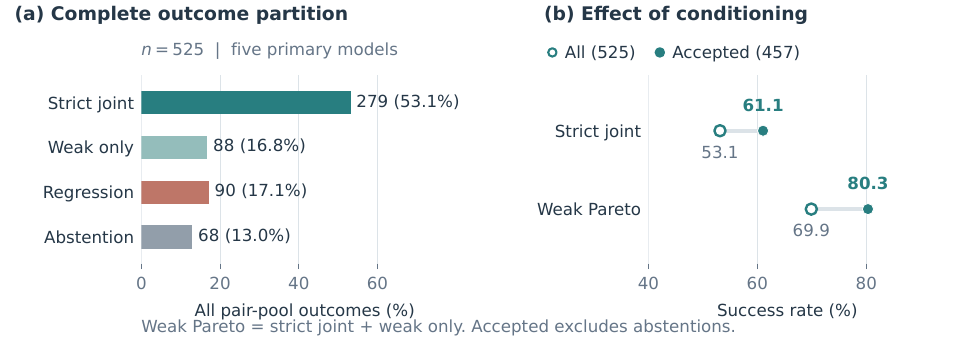}
\caption{\textbf{Abstention is part of the deployed outcome.} All 525 primary evaluations partition into 279 strict successes, 88 weak-only improvements, 90 regressions, and 68 Base fallbacks. Weak Pareto includes the first two categories. Any conditional rates exclude the 68 fallbacks and use denominator 457; primary comparisons always retain denominator 525.}
\label{fig:outcomes}
\end{figure}
\paragraph{Abstention returns Base, while accepted merges can still regress.}
Weak Pareto covers $367/525=69.9\%$ of evaluations; $90/525=17.1\%$ regress and $68/525=13.0\%$ abstain (Figure~\ref{fig:outcomes}). Abstentions contribute zero to WG and mean gain, which are $+0.61$ and $+1.57$ pp. Among the 457 non-abstaining evaluations, SJ is $61.1\%$; this conditional rate is not substituted for the unconditional $53.1\%$. Positive validation gains therefore do not eliminate held-out regressions.

\begin{figure}[t]
\centering
\includegraphics[width=\linewidth]{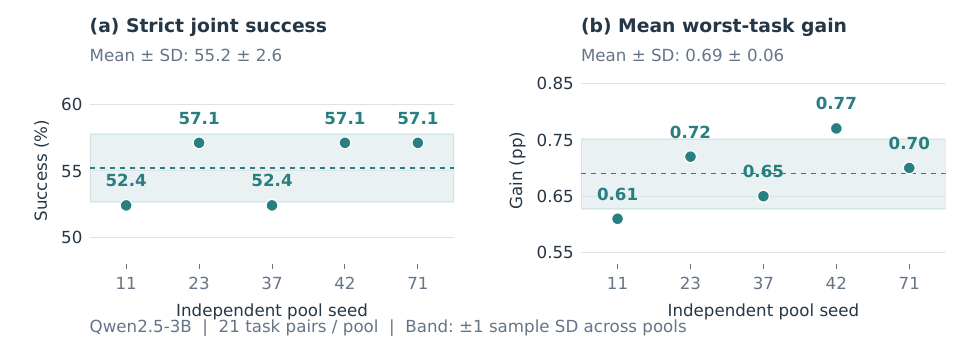}
\caption{\textbf{Joint gains persist across five Qwen2.5-3B pools.} Each pool evaluates the same 21 task pairs under the fixed protocol. Points show the individual pools; summaries are mean $\pm$ sample SD, not confidence intervals. Pool-level variation addresses candidate-search randomness and does not quantify all test-sample uncertainty.}
\label{fig:pools}
\end{figure}
\begin{figure}[t]
\centering
\includegraphics[width=\linewidth]{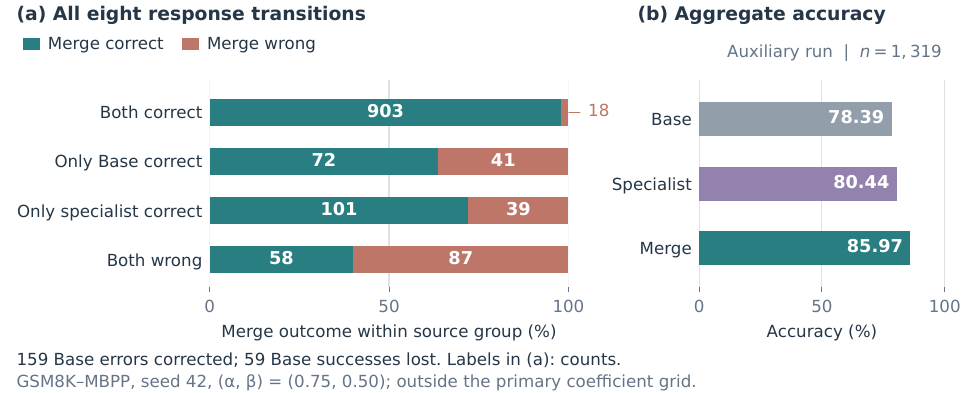}
\caption{\textbf{Corrections and losses in an auxiliary coefficient analysis.} Qwen2.5-3B, GSM8K--MBPP, pool 42; all 1,319 GSM8K test items. The selected pair is additionally evaluated at $(\alpha_{\rm GSM},\beta_{\rm MBPP})=(0.75,0.50)$, outside the five-point primary grid. This run contributes to neither main table. Specialist denotes the GSM8K expert participating in this merge.}
\label{fig:transitions}
\end{figure}

\subsection{Matched pair-ranking controls (RQ3)}
\label{sec:rq3}
\paragraph{Functional quality, global geometry, and layerwise subspaces.}
Quality-only ranks functional evidence; Global Cosine supplements it with global vector geometry, Ours with layerwise singular-subspace geometry. Other pipeline settings remain fixed. SJ increases $49.1<50.5<53.1\%$, WG increases $0.49<0.54<0.61$ pp, Weak Pareto rises $66.1\to67.6\to69.9\%$, and mean gain rises $1.39\to1.48\to1.57$ pp; regression decreases $20.4>19.0>17.1\%$ and abstention decreases $13.5\to13.3\to13.0\%$. Directionally consistent aggregate retention and deployment metrics support incremental utility of layerwise ranking within this matched progression, without implying statistical significance or a causal mechanism.

Improved retention is not explained by more fallback. With the same pools and at most 50 proposals per task pair, overlap changes which combinations reach validation. Appendix~\ref{app:additional} adds matched configuration-level and one-factor-at-a-time robustness analyses.

\begin{samepage}
\subsection{Independent-pool variation (RQ4)}
\label{sec:rq4}
For seeds 11, 23, 37, 42, and 71, Qwen 3B SJ ranges from $52.4\%$ to $57.1\%$, averaging $55.2\pm2.6\%$; WG ranges from $+0.61$ to $+0.77$ pp, averaging $+0.69\pm0.06$ pp (Figure~\ref{fig:pools}). All five pools have positive WG. These pool-level summaries describe search variability rather than inferential uncertainty across ranking methods; Appendix~\ref{app:scopes} discusses the dependence structure of the reported evaluations.

\end{samepage}

\subsection{Auxiliary response analysis}
\label{sec:auxiliary}
\paragraph{Corrections outweigh losses in this auxiliary run.}
Base, the participating GSM8K specialist, and Merge answer 1,034, 1,061, and 1,134 items correctly ($78.39\%$, $80.44\%$, $85.97\%$). Merge corrects 159 Base errors and loses 59 Base-correct answers, giving $+100/1{,}319=7.58$ pp. It corrects 58 errors shared by the two displayed references, but also loses 18 answers both answered correctly (Figure~\ref{fig:transitions}). These counts describe auxiliary behavior, including failures; they establish neither primary-grid correction rates nor a capability absent from both constituent experts.
\begin{figure}[t]
\centering
\includegraphics[width=\linewidth]{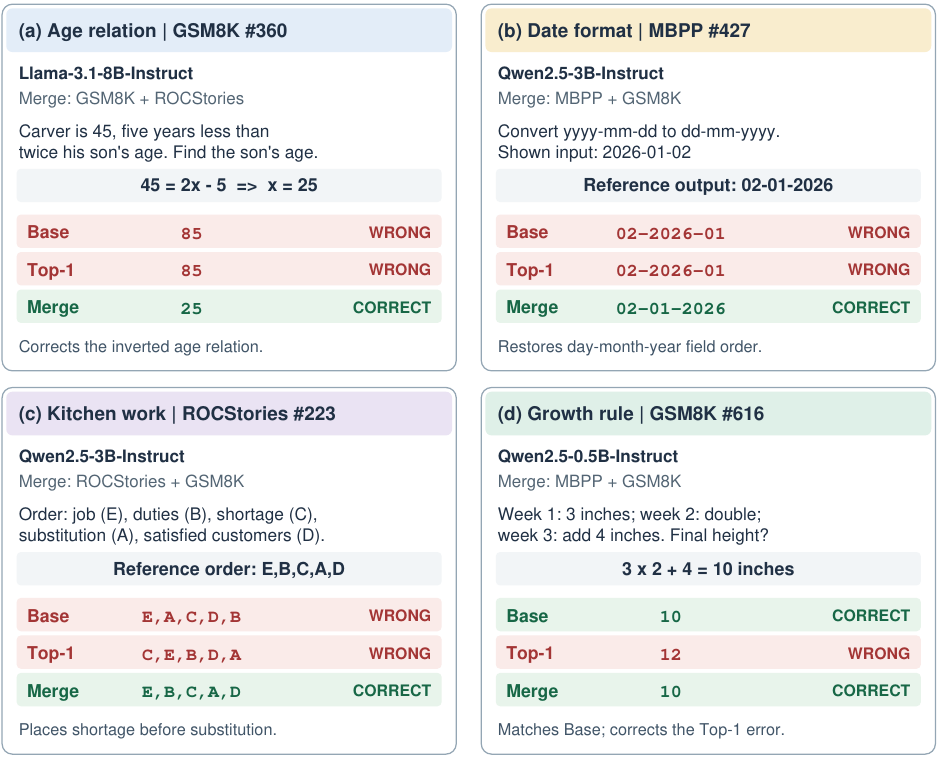}
\caption{\textbf{Selected qualitative examples of equal-weight merging.} S1--S3 correct displayed shared errors; S4 preserves Base correctness. These separately collected $0.5/0.5$ examples illustrate behaviors and are distinct from both the primary comparison and the auxiliary coefficient run in Figure~\ref{fig:transitions}.}
\label{fig:cases}
\end{figure}
\paragraph{Qualitative examples illustrate forms of recovery.}
\label{sec:cases}
Figure~\ref{fig:cases} illustrates three shared-error corrections (S1--S3) and one Base-preserving merge (S4). S1 solves $45=2x-5$ as $x=25$, where both references answer 85. S2 converts \texttt{2026-01-02} to \texttt{02-01-2026} instead of \texttt{02-2026-01}; executing the supplied code gives Merge $3/3$ passed tests versus $0/3$ for both references. S3 restores employment $\to$ duties $\to$ shortage $\to$ substitution $\to$ customer outcome.

S4 preserves Base's $3\times2+4=10$ versus the specialist's 12; this is the merged response, not Base fallback. These selected equal-weight examples illustrate behavior, not the full spectral selector or prevalence. Appendix~\ref{app:cases} preserves full responses and provenance.

\section{Conclusion}
Local expert composition is a joint-retention pair-selection problem: shared origin, proximity, and orthogonality do not guarantee success. Quality-constrained spectral selection combines functional screening, pair geometry, calibration, and abstention under a fixed validation budget. Matched controls support incremental spectral utility, with persistent regressions and abstentions. Stronger predictive accounts, radius dependence, and larger expert sets remain open. By separating expert quality, pair compatibility, validation, and abstention within a joint-retention framework, the study makes local expert composition an explicit selection problem.

\clearpage
\section*{AI Use Statement}
In this work, we used generative AI tools for limited assistance with technical tasks. We have not used generative AI tools for other applicable tasks with required disclosure, and the remaining required-disclosure tasks are not applicable to this work. Additionally, we used generative AI tools to edit the manuscript to improve readability. We have reviewed all AI-assisted work. All AI-generated suggestions were independently reviewed, revised, and validated by the authors before inclusion in the final work. We take responsibility for the final content of this work, including text, claims, or artifacts produced with the aid of generative AI.
\section*{Ethics Statement}
This work uses existing model checkpoints and standard research benchmarks and does not involve the collection of private data or human-subject experiments. We follow the applicable licenses and usage conditions of the models, datasets, and software used in this study.
\section*{Reproducibility Statement}
We provide the methodological details, experimental settings, evaluation protocols, and hyperparameters needed to understand and reproduce the reported experiments in the main paper and appendices. An anonymous code repository with the corresponding implementation and supporting materials is also provided.
\bibliography{references}
\bibliographystyle{iclr2027_conference}
\clearpage
\appendix
\section{Elementary facts about local composition}
\label{app:theory}

These elementary identities and counterexamples clarify the limits of parameter
proximity and orthogonality. The loss bounds concern smooth surrogates and do not
guarantee discrete benchmark accuracy gains.
Let $\theta_0\in\mathbb{R}^{d}$ be a shared checkpoint and let $u,v\in\mathbb{R}^{d}$
be the two updates actually added to it. Coefficients can be absorbed into these
updates: for equal-weight averaging, for example, use
$u=\delta_A/2$ and $v=\delta_B/2$.

\subsection{An exact interaction identity}
\label{app:interaction_identity}

For task $t$, define the individual loss change
$\Delta L_t(w)=L_t(\theta_0+w)-L_t(\theta_0)$ and the interaction
\begin{equation}
 I_t(u,v)=L_t(\theta_0+u+v)-L_t(\theta_0+u)
          -L_t(\theta_0+v)+L_t(\theta_0).
 \label{eq:app_interaction}
\end{equation}
Then, without smoothness assumptions,
\begin{equation}
 \Delta L_t(u+v)=\Delta L_t(u)+\Delta L_t(v)+I_t(u,v).
 \label{eq:app_decomposition}
\end{equation}
If $L_t$ is twice continuously differentiable on an open set containing
$\{\theta_0+s u+r v:s,r\in[0,1]\}$, two applications of the fundamental theorem
of calculus give the exact formula
\begin{equation}
 I_t(u,v)=\int_0^1\!\int_0^1
 u^{\mathsf T}\nabla^2 L_t(\theta_0+s u+r v)v\,dr\,ds.
 \label{eq:app_hessian_integral}
\end{equation}
Indeed, apply the theorem to the mixed derivative of
$f(s,r)=L_t(\theta_0+s u+r v)$.
If the Hessian operator norm is bounded by $K_t$ throughout this parallelogram,
\begin{equation}
 |I_t(u,v)|\leq K_t\lVert u\rVert_2\lVert v\rVert_2.
 \label{eq:app_interaction_bound}
\end{equation}
Euclidean orthogonality does not generally annul this Hessian-weighted interaction.

\subsection{A sufficient condition for joint loss improvement}
\label{app:sufficient_condition}

For every target task $t$, suppose there is a loss margin $\mu_t>0$ such that
\begin{equation}
 \Delta L_t(u)+\Delta L_t(v)\leq -\mu_t,
 \qquad K_t\lVert u\rVert_2\lVert v\rVert_2<\mu_t.
 \label{eq:app_joint_condition}
\end{equation}
Equations~\eqref{eq:app_decomposition}--\eqref{eq:app_interaction_bound} imply
$\Delta L_t(u+v)<0$ for every task. A directly established upper bound on
$I_t(u,v)$ can replace the Hessian bound.
The first condition involves \emph{both} updates on \emph{each} task;
improvement of each specialist on its own task is insufficient.
The condition separates the isolated benefits from the interaction that they
must overcome. Its bounds and margins have not been established for the LLMs
in the reported evaluations.

\subsection{Orthogonal specialists can yield a harmful merge}
\label{app:orthogonal_counterexample}

Consider higher-is-better scores
\begin{equation}
 g_A(x,y)=x-2y,\qquad g_B(x,y)=y-2x
\end{equation}
at the base $(0,0)$, with specialists $a=(\varepsilon,0)$ and
$b=(0,\varepsilon)$ for any $\varepsilon>0$.
Their updates are orthogonal and have norm $\varepsilon$, while
$g_A(a)=g_B(b)=\varepsilon>0$.
Nevertheless,
\begin{equation}
 g_A(a+b)=g_B(a+b)=-\varepsilon,
 \qquad
 g_A\!\left(\frac{a+b}{2}\right)
 =g_B\!\left(\frac{a+b}{2}\right)=-\frac{\varepsilon}{2}.
\end{equation}
Thus addition and equal-weight averaging both underperform the base on both
tasks, however small the neighborhood. Taking $L_t=-g_t$ yields smooth losses
with zero interaction: the failure comes entirely from each specialist's
cross-task harm. This counterexample refutes an implication from proximity
and orthogonality alone; it does not model benchmark accuracy.

\subsection{Two-vector Gram spectra contain no extra information}
\label{app:gram_spectrum}

For nonzero flattened updates, write $\hat u=u/\lVert u\rVert_2$,
$\hat v=v/\lVert v\rVert_2$, and $c=\hat u^{\mathsf T}\hat v$.
The normalized Gram matrix and its ordered eigenvalues are
\begin{equation}
 G=\begin{pmatrix}1&c\\c&1\end{pmatrix},\qquad
 \lambda_{\max}(G)=1+|c|,\quad
 \lambda_{\min}(G)=1-|c|.
\end{equation}
Consequently, a statistic using only this spectrum cannot add information
beyond absolute cosine similarity and also discards its sign. This conclusion
does not extend to arbitrary layerwise singular-subspace statistics.

\subsection{Merging changes the perturbation radius}
\label{app:radius_identity}

For arbitrary real coefficients $\alpha,\beta$,
\begin{equation}
 \lVert\alpha u+\beta v\rVert_2^2
 =\alpha^2\lVert u\rVert_2^2+\beta^2\lVert v\rVert_2^2
   +2\alpha\beta\,u^{\mathsf T}v.
\end{equation}
If both updates have norm $\rho$ and signed cosine $c$, their average has
norm $\rho\sqrt{(1+c)/2}$ and their sum has norm
$\rho\sqrt{2+2c}$. At $c=0$, these are $\rho/\sqrt{2}$ and
$\sqrt{2}\rho$, respectively. Geometric selection must therefore be assessed
alongside radius controls. For a nonlinear merger such as TIES, these identities
apply only to an explicit additive decomposition of its actual transformed
output, not automatically to the raw expert updates.

\section{Experimental Protocol and Result Provenance}
\label{app:protocol}
\subsection{Evaluation scope and fixed configuration}
Each model is evaluated on all 21 unordered pairs of seven tasks, using five independently sampled perturbation pools (seeds 11, 23, 37, 42, and 71). Table~\ref{tab:results} therefore contains 105 pair--pool evaluations per model and method. The primary aggregate in Table~\ref{tab:methods} covers Qwen2.5-1.5B-Instruct, Qwen2.5-3B-Instruct, Qwen2.5-7B-Instruct, Llama-3.1-8B-Instruct, and OLMo3-7B-Instruct: 525 evaluations per method. Qwen2.5-0.5B-Instruct remains a separate scale stress test. Equal sample counts give each model in this aggregate equal weight. In this study, ``nearby'' refers operationally to experts drawn from the recorded local perturbation pools around the same shared checkpoint. We do not use the term to imply a universal absolute parameter-space radius.

\begin{table}[htbp]
\centering\small
\setlength{\tabcolsep}{6pt}\renewcommand{\arraystretch}{1.10}
\caption{\textbf{Fixed primary configuration.} Settings apply across model families, task pairs, and pools. Gains and thresholds use percentage points (pp).}\label{tab:protocol}
\begin{tabular}{@{}p{.36\linewidth}p{.59\linewidth}@{}}\toprule
\rowcolor{tablehead}Component & Setting\\\midrule
Candidate population & $N=5{,}000$ per independent pool\\
\rowcolor{tablerow}Retained experts & Up to $K=50$ per task; $g_t^{\mathrm{sel}}\geq0$\\
Pair shortlist & Up to $B=10$ distinct expert pairs\\
\rowcolor{tablerow}Coefficient grid $\mathcal A$ & $(0.5,0.5)$, $(0.25,0.75)$, $(0.75,0.25)$, $(1,0.5)$, $(0.5,1)$\\
Merge validation budget & At most $10\times5=50$ proposals per task pair\\
\rowcolor{tablerow}Spectral coefficient & $\lambda=0.5$ for every primary configuration\\
Spectral rank & Cap 8; common positive numerical rank per compared tensor\\
\rowcolor{tablerow}Validation gate & $g_A^{\mathrm{val}}>0$ and $g_B^{\mathrm{val}}>0$\\
Selection / validation data & 200 / 200 examples per task; disjoint splits\\
\rowcolor{tablerow}Final assessment & Official held-out test split; excluded from selection\\
Inference & BF16; greedy decoding; one response per item\\
\rowcolor{tablerow}Deployed model & One accepted merge, or Base on abstention\\
\bottomrule\end{tabular}
\end{table}

The same primary configuration is applied uniformly across models, task pairs, and perturbation pools; the study evaluates this fixed procedure rather than claiming that these settings are universally optimal.

\paragraph{Candidate-pool provenance.}
All reported evaluations use fixed realized perturbation pools generated for the experiments and identified by model, pool size, and seed. Candidate scoring, expert selection, merging, validation, and held-out evaluation were performed on these realized pools, and Tables~\ref{tab:results}--\ref{tab:methods} report outcomes from those completed runs. We therefore treat the realized pools as the experimental sampling units for the reported analyses and make no empirical claim about performance as a function of perturbation radius.

Candidate scoring, screening, and pair ranking use only selection examples. Coefficient selection and the joint-positive gate use only validation examples. For MATH-500, both development subsets are drawn from MATH training data and MATH-500 is reserved for final assessment. Validation ties do not pass the gate. Among feasible proposals, selection maximizes worst-task gain, then mean gain, then fixed candidate/coefficient order; an empty feasible set returns Base. The validation split is used for bounded proposal selection among at most 50 candidate merges, while the final reported gains are computed only on an untouched held-out test split.

\subsection{Aggregation, abstention, and uncertainty}
\label{app:scopes}
Let $g_A,g_B$ denote test gains relative to Base. Strict joint improvement requires $\min(g_A,g_B)>0$. Weak Pareto improvement requires $\min(g_A,g_B)\geq0$ and $\max(g_A,g_B)>0$, and thus includes strict joint improvement. Regression requires $\min(g_A,g_B)<0$. Abstention is the selection decision to return Base; it contributes zero to both gains and belongs to none of these improvement or regression categories. Every unconditional rate and gain average includes abstentions in its denominator. Worst-task gain averages $\min(g_A,g_B)$ across records; mean gain averages $(g_A+g_B)/2$.

The confirmed primary outcome counts partition all 525 evaluations into strict improvement, weak-only improvement, regression, and abstention. An accepted merge that ties Base on both test tasks would require a separate category; the confirmed primary counts contain no such records. Conditional rates, when shown, divide by the 457 non-abstaining evaluations and are explicitly labeled. This conditional view does not replace unconditional reporting.

Five-pool summaries describe search variability. Sample standard deviation across the five Qwen2.5-3B-Instruct pools is descriptive, not a confidence interval. Shared assessment items and overlapping task pairs prevent treating all pair--pool records as independent trials. Additional matched configuration-level outcome counts are reported in Appendix~\ref{app:additional}; we use these paired records descriptively rather than reconstructing inferential statistics from rounded aggregate summaries.

\subsection{Controls and auxiliary response analysis}
Quality-only selection is exactly the $\lambda=0$ control of Eq.~\eqref{eq:ranking}; screening, coefficients, validation data, and the gate remain fixed. Writing its quality term as $Q(i,j)$, the corresponding global-cosine control is
\[
J_{\mathrm{cos}}(i,j)=Q(i,j)-\lambda
\left|\frac{\langle\delta_i,\delta_j\rangle}{\|\delta_i\|_2\|\delta_j\|_2}\right|,
\]
where $\delta_i,\delta_j$ are flattened full updates. Quality-only, Global Cosine, and Ours share the same candidate pools, screening rule, coefficient set, validation budget, and decision gate; they differ only in pair ranking. Both controls use the same 525 evaluation configurations as Ours. Table~\ref{tab:methods} reports their aggregate outcomes. These are component comparisons of the pair-ranking criterion; Top-1, TIES, and Direct Search instead compare complete merge/search systems. TIES is included as a standard interference-handling merge comparator and is interpreted as a full-system baseline rather than a matched ablation of the spectral ranking term.

The auxiliary run uses the selected GSM8K--MBPP pair from Qwen2.5-3B-Instruct, pool 42, with $N=5{,}000$ and $K=50$. All 1,319 GSM8K test items use $(\alpha,\beta)=(0.75,0.50)$, BF16 greedy decoding, and one response per item. These coefficients lie outside $\mathcal A$; the run enters neither primary table nor pool averages. ``Specialist'' is the participating GSM8K expert. Eight count-derived categories exhaust this run; separately collected qualitative cases do not estimate population frequencies.

\paragraph{Analysis granularity.}
The archived experiment summaries preserve the reported model-level and primary aggregate results used in Tables~\ref{tab:results}--\ref{tab:methods}, together with the selected response analyses. Additional matched configuration-level outcome counts and one-factor-at-a-time sensitivity analyses are now reported in Appendix~\ref{app:additional}; their primary-configuration values agree with Table~\ref{tab:methods}. Proposal-level validation traces and full pairwise geometry records are not used to derive the reported primary results or the matched analyses in Appendix~\ref{app:additional}. The available matched counts are reported descriptively, and we do not reconstruct inferential statistics from rounded aggregate summaries.

\clearpage
\subsection{Selection pseudocode}
\label{app:algorithm}
Sampling precedes Algorithm~\ref{alg:selection}. Primary margins are $\eta_t=m_t=0$: screening is non-strict and validation is strict.

\begin{algorithm}[H]
\caption{Quality-constrained spectral selection}
\label{alg:selection}
\small
\begin{algorithmic}[1]
\Require $\base$, task-scored candidates, separate validation data;
\Statex \hspace{1.5em} $K,B,\eta_t,\lambda,r_\ell,w_\ell,\mathcal{A}$ and validation margins $m_t$.
\State $\mathcal{E}_t\gets$ up to $K$ best candidates with $g_t^{\mathrm{sel}}\geq\eta_t$, for $t=A,B$.
\State $\mathcal{P}\gets\{(i,j):i\in\mathcal{E}_A,\ j\in\mathcal{E}_B,\ i\neq j\}$.
\State Rank $\mathcal{P}$ by $J(i,j)$; retain up to $B$ pairs; set $\mathcal{F}\gets\varnothing$.
\For{each retained pair and each $(\alpha,\beta)\in\mathcal{A}$}
\State Construct $\theta$ using Eq.~\eqref{eq:merge}; evaluate $g_A^{\mathrm{val}},g_B^{\mathrm{val}}$.
\If{$g_t^{\mathrm{val}}(\theta)>m_t$ for both tasks}
\State Add $\theta$ and its candidate/coefficient identifiers to $\mathcal{F}$.
\EndIf
\EndFor
\If{$\mathcal{F}=\varnothing$}
\State \Return $\base$; record abstention.
\EndIf
\State \Return the feasible $\theta\in\mathcal{F}$ maximizing worst-task validation gain, breaking ties by mean validation gain and then by fixed candidate/coefficient order.
\end{algorithmic}
\end{algorithm}

\clearpage
\section{Case Selection, Responses, and Provenance}
\label{app:cases}
\paragraph{Collection and selection.}
The qualitative archive contains 4,945 response records collected for case analysis. Its recorded categories comprise 1,958 shared-error corrections, 2,226 Base-preserving recoveries, and 761 additional records not used as case studies. Because this archive was constructed for qualitative case discovery rather than prevalence estimation, these category counts are not used as quantitative evaluation denominators or overall improvement rates. The five examples below are the accompanying case file's recommended examples. S1--S4 appear in Figure~\ref{fig:cases}(a)--(d), respectively; S5 provides an additional instance of Base retention.

Per-example responses were collected separately using saved expert configurations. We therefore treat these examples as observed response comparisons and keep them separate from the quantitative results reported in Tables~\ref{tab:results}--\ref{tab:methods}. All five use equal-weight perturbation merging. Only the assessment-task Top-1 response is displayed; the other constituent expert is not shown. The cases therefore illustrate merging behavior rather than establishing a capability absent from both constituent experts. These examples illustrate equal-weight merging rather than the full spectral selector. The separate transition analysis in Figure~\ref{fig:transitions} quantifies successes and failures over 1,319 responses from a separate auxiliary coefficient analysis.

\begin{table}[h]
\centering
\small
\setlength{\tabcolsep}{4pt}
\renewcommand{\arraystretch}{1.15}
\caption{\textbf{Case identities and observed answers.} Test indices are zero-based within the supplied archive; task and story identifiers are retained as recorded. Qwen denotes Qwen2.5-Instruct; Llama denotes Llama-3.1-8B-Instruct. S2 shows outputs from executing the generated code on the first displayed test, rather than literal model replies. Letter strings preserve the complete story order.}
\label{tab:case-identities}
\begin{tabular}{@{}l>{\raggedright\arraybackslash}p{0.37\linewidth}lll@{}}
\toprule
\rowcolor{tablehead}
Case & Model, merge pair, and item & Base & Top-1 & Merge\\
\midrule
S1 & Llama; GSM8K + ROCStories; GSM8K test 360 & 85 & 85 & \textbf{25}\\
\rowcolor{tablerow}
S2 & Qwen 3B; MBPP + GSM8K; MBPP task 427, test 680 & \multicolumn{3}{l}{Date conversion; outputs below}\\
 & Input: \texttt{2026-01-02} & \texttt{02-2026-01} & \texttt{02-2026-01} & \textbf{\texttt{02-01-2026}}\\
\rowcolor{tablerow}
S3 & Qwen 3B; ROCStories + GSM8K; story/test 223 & \texttt{EACDB} & \texttt{CEBDA} & \textbf{\texttt{EBCAD}}\\
S4 & Qwen 0.5B; MBPP + GSM8K; GSM8K test 616 & \textbf{10} & 12 & \textbf{10}\\
\rowcolor{tablerow}
S5 & Qwen 3B; ROCStories + GSM8K; story/test 334 & \textbf{\texttt{ABDEC}} & \texttt{ADEBC} & \textbf{\texttt{ABDEC}}\\
\bottomrule
\end{tabular}
\end{table}

\paragraph{Response-level checks.}
The displayed age and beanstalk answers follow directly from $45=2x-5$ and $3\times2+4$, respectively. In S2, the shared reference implementation uses the substitution \texttt{\textbackslash3-\textbackslash1-\textbackslash2}; the merged implementation extracts year, month, and day and returns them in reverse order. Executing the supplied code on the three displayed tests gives Base $0/3$, Top-1 $0/3$, and Merge $3/3$. These tests cover \texttt{2026-01-02}, \texttt{2020-11-13}, and \texttt{2021-04-26}; they do not establish behavior on other inputs. In S3, the reference order is kitchen employment (E), duties (B), shortage (C), substitution (A), and customer outcome (D). In S5, fire construction (A) precedes loss of control (B), fetching the hose (D), containment (E), and final extinguishing (C). Top-1 places containment before loss of control; Base and Merge match the reference order.

\paragraph{Traceability.}
The case-analysis package retains the case descriptions and a structured selection with exact prompts, reference answers, full responses, and source-record identifiers. S1 belongs to experiment group \texttt{6d04bf98cd172ce0}; S2 to \texttt{3a2b2d0bb0bf16cf}; S3 and S5 to \texttt{bb35fcbb12a89ce2}; and S4 to \texttt{c454bac537eef084}. The recorded experiment-group identifiers and source paths provide provenance for the displayed cases. Figure~\ref{fig:cases} uses shortened prompts and response summaries for readability; full wording remains in the accompanying records.

\clearpage
\section{Additional Matched and Robustness Analyses}
\label{app:additional}
\renewcommand{\thetable}{D.\arabic{table}}
\renewcommand{\theHtable}{D.\arabic{table}}
\setcounter{table}{0}

\subsection{Configuration-level matched analysis}
\label{app:matched}
Because Quality-only, Global Cosine, and the spectral selector operate on the same 525 model--task-pair--pool configurations, we additionally examine their strict-joint outcomes in matched form. This analysis complements the aggregate results in Table~\ref{tab:methods} by showing how often spectral ranking changes the outcome of the same evaluation configuration. All counts use the same unconditional denominator of 525, including configurations that ultimately return Base.

\begin{table}[htbp]
\centering\footnotesize
\setlength{\tabcolsep}{3pt}\renewcommand{\arraystretch}{1.18}
\caption{Configuration-level paired comparison of strict joint success. Each row partitions all 525 matched model--task-pair--pool configurations. ``Ours only'' denotes a strict joint success for the spectral selector when the comparator fails; ``Comparator only'' denotes the reverse. Delta SJ is computed from the underlying counts.}
\label{tab:paired}
\begin{tabular}{@{}>{\raggedright\arraybackslash}p{.26\linewidth}*{6}{>{\raggedleft\arraybackslash}p{\dimexpr(.74\linewidth-12\tabcolsep)/6\relax}}@{}}\toprule
\rowcolor{tablehead}Comparison & \shortstack[r]{Both\\succeed} & \shortstack[r]{Ours only\\succeeds} & \shortstack[r]{Comparator\\only\\succeeds} & \shortstack[r]{Both\\fail} & \shortstack[r]{$\Delta$ SJ\\(pp)} & \shortstack[r]{$\Delta$ WG\\(pp)}\\\midrule
Ours vs. Quality-only & 246 & \textbf{33} & 12 & 234 & \textbf{+4.0} & \textbf{+0.12}\\
\rowcolor{tablerow}Ours vs. Global Cosine & 251 & \textbf{28} & 14 & 232 & \textbf{+2.7} & \textbf{+0.07}\\
\bottomrule\end{tabular}
\end{table}

Against Quality-only, spectral ranking yields 33 Ours-only strict successes versus 12 comparator-only successes. Against Global Cosine, the corresponding counts are 28 versus 14. Thus, the aggregate improvement is associated with more favorable than unfavorable configuration-level outcome changes. The exact count-based SJ differences are $+4.0$ pp over Quality-only and $+2.7$ pp over Global Cosine.

\begin{table}[htbp]
\centering\footnotesize
\setlength{\tabcolsep}{3pt}\renewcommand{\arraystretch}{1.18}
\caption{Model-wise strict joint success for the three matched pair-ranking variants. Each model contains 21 task pairs and five perturbation pools, for 105 configurations. The final column reports the count-based difference between Ours and Global Cosine.}
\label{tab:matched_models}
\begin{tabular}{@{}>{\raggedright\arraybackslash}p{.22\linewidth}*{3}{>{\raggedleft\arraybackslash}p{\dimexpr.20\linewidth-2\tabcolsep\relax}}>{\raggedleft\arraybackslash}p{\dimexpr.18\linewidth-2\tabcolsep\relax}@{}}\toprule
\rowcolor{tablehead}Model & \shortstack[r]{Quality-only\\SJ} & \shortstack[r]{Global Cosine\\SJ} & Ours SJ & \shortstack[r]{Ours $-$ Cosine\\(pp)}\\\midrule
Qwen2.5-1.5B & $40/105=38.1\%$ & $42/105=40.0\%$ & $45/105=42.9\%$ & +2.9\\
\rowcolor{tablerow}Qwen2.5-3B & $53/105=50.5\%$ & $55/105=52.4\%$ & $58/105=55.2\%$ & +2.9\\
Qwen2.5-7B & $57/105=54.3\%$ & $59/105=56.2\%$ & $62/105=59.0\%$ & +2.9\\
\rowcolor{tablerow}Llama-3.1-8B & $54/105=51.4\%$ & $55/105=52.4\%$ & $60/105=57.1\%$ & +4.8\\
OLMo 3-7B & $54/105=51.4\%$ & $54/105=51.4\%$ & $54/105=51.4\%$ & 0.0\\\midrule
\rowcolor{tablehead}\textbf{Total} & $258/525=49.1\%$ & $265/525=50.5\%$ & $279/525=53.1\%$ & +2.7\\
\bottomrule\end{tabular}
\end{table}

The improvement over Global Cosine is not concentrated in a single checkpoint. Spectral ranking improves strict joint success on four of the five primary checkpoints and ties on OLMo 3-7B. The four positive model-wise differences range from $+2.9$ to $+4.8$ pp.

\clearpage
\subsection{One-factor-at-a-time robustness}
\label{app:robustness}
We next vary one primary design choice at a time while holding the remaining pipeline settings fixed. These experiments assess whether the primary result depends on a narrow choice of spectral weight, subspace rank, pair-validation budget, or retained candidate-pool size. The default configuration is $\lambda=0.5$, rank cap $=8$, $B=10$, and $K=50$.

\begin{table}[htbp]
\centering\small
\setlength{\tabcolsep}{6pt}\renewcommand{\arraystretch}{1.18}
\caption{One-factor-at-a-time robustness analysis. Unless the indicated factor is changed, all remaining settings are held at the primary configuration. SJ and Regression are percentages; WG is in percentage points.}
\label{tab:robustness}
\begin{tabular}{@{}>{\raggedright\arraybackslash}p{.18\linewidth}>{\raggedright\arraybackslash}p{.28\linewidth}*{3}{>{\raggedleft\arraybackslash}p{\dimexpr(.54\linewidth-8\tabcolsep)/3\relax}}@{}}\toprule
\rowcolor{tablehead}Factor & Setting & \shortstack[r]{SJ\\(\%)} & \shortstack[r]{WG\\(pp)} & \shortstack[r]{Regression\\(\%)}\\\midrule
$\lambda$ & $0$ (Quality-only) & 49.1 & +0.49 & 20.4\\
$\lambda$ & 0.25 & 51.8 & +0.56 & 18.7\\
\rowcolor{tablehead}$\lambda$ & \textbf{0.50} & 53.1 & +0.61 & 17.1\\
$\lambda$ & 0.75 & 52.8 & +0.60 & 17.5\\\midrule
Rank cap & 4 & 52.4 & +0.58 & 17.9\\
\rowcolor{tablehead}Rank cap & \textbf{8} & 53.1 & +0.61 & 17.1\\
Rank cap & 16 & 52.8 & +0.60 & 17.5\\\midrule
$B$ & 5 & 51.6 & +0.56 & 18.5\\
\rowcolor{tablehead}$B$ & \textbf{10} & 53.1 & +0.61 & 17.1\\
$B$ & 20 & 53.5 & +0.62 & 17.0\\\midrule
$K$ & 25 & 51.8 & +0.56 & 18.3\\
\rowcolor{tablehead}$K$ & \textbf{50} & 53.1 & +0.61 & 17.1\\
$K$ & 100 & 53.0 & +0.60 & 17.3\\
\bottomrule\end{tabular}
\end{table}

The sensitivity results show a broad operating region rather than a narrow optimum. Introducing a moderate spectral penalty improves SJ relative to the $\lambda=0$ quality-only control, while $\lambda$ values from 0.25 to 0.75 remain close to the primary result. Performance is similarly stable for rank caps from 4 to 16. Increasing the pair shortlist from $B=5$ to $B=10$ improves retention, whereas $B=20$ provides only a small additional gain, indicating diminishing returns beyond the primary validation budget. Increasing $K$ from 25 to 50 is beneficial, while $K=100$ yields essentially no further improvement. Overall, the primary configuration lies in a stable region rather than at an isolated favorable setting.

\end{document}